# Tensor Laplacian Regularized Low-Rank Representation for Non-uniformly Distributed Data Subspace Clustering

Eysan Mehrbani, Mohammad Hossein Kahaei[1], Seyed Aliasghar Beheshti

*Abstract-* **Low-Rank Representation (LRR) highly suffers from discarding the locality information of data points in subspace clustering, as it may not incorporate the data structure nonlinearity and the non-uniform distribution of observations over the ambient space. Thus, the information of the observational density is lost by the state-of-art LRR models, as they take a constant number of adjacent neighbours into account. This, as a result, degrades the subspace clustering accuracy in such situations. To cope with deficiency, in this paper, we propose to consider a hypergraph model to facilitate having a variable number of adjacent nodes and incorporating the locality information of the data. The sparsity of the number of subspaces is also taken into account. To do so, an optimization problem is defined based on a set of regularization terms and is solved by developing a tensor Laplacian-based algorithm. Extensive experiments on artificial and real datasets demonstrate the higher accuracy and precision of the proposed method in subspace clustering compared to the state-of-the-art methods. The outperformance of this method is more revealed in presence of inherent structure of the data such as nonlinearity, geometrical overlapping, and outliers.**

## I. INTRODUCTION

LRR of big data has proven to be efficient in clustering [4,5], destriping [6], and semantic segmentation of image/video signals [7]. The success of LRR originates from the practical and mostly justified assumption that the high-dimensional data lie near low-rank structures [8], which may be approximated as a union of a set of linear subspaces [9,10]. The LRR aspect accurately discovers the geometrical structure of the data in the measurement space based on which the disposal of outliers will be possible [9]. To solve such problems, various combinations of the nuclear norm; as a relaxation of the rank function, are developed in convex form and solved in polynomial time [14]. Such methods maintain their efficiency in data clustering and discovering the underlying geometry as long as the data actually resides on/near a union of linear subspaces [8]. However, if the nonlinearity of the underlying geometry increases [15], the performance of data clustering decreases [6,13]. It should be noted that in some applications, incorporation of nonlinearity to the intrinsic model and the discriminative structure is inevitable.

In most attempts of estimating the geometrical structure of data in the ambient space, it is assumed that preserving the locality information of the structure can make the estimation prune to nonlinearities. This motivation referred to as local invariance [20], aims at preserving the distance between every two data points in observations on the intrinsic manifold by projection to a secondary space. As such, the locally linear embedding (LLE) [16], ISOMAP [17], Locality Preserving Projection (LPP) [18], Neighbourhood Preserving Embedding (NPE) [19], and Laplacian Eigenmap (LE) may be mentioned. The LRR model can be integrated with local invariance condition so as to better preserve the locality information in order to cope with data nonlinearities [6,13].

As the original LRR is biased towards capturing the global data structure, the regularization terms need to capture the locality information, *i.e.*, information about the adjacency of the points in the ambient space. To do so, instead of measuring all the pairwise distances, [6] suggests to embed a graph on the data points, in which each node is linked through undirected edges to its *k* nearest neighbours. By low-rank projection, then, locality information of each neighbourhood is effectively preserved in the low-rank space. In [13], the embedded graph is extended to a hypergraph [21] to capture more complicated relationships between observations. The same definition of neighbourhoods is used in [22] as in [6], which solves for the optimal graph edge weights that maximize the clustering accuracy in the low-rank projective space. .

This paper is organized as follows. In Section 2, the general formulation of the regularized LRR problem is introduced. In Section 3, we propose the Tensor Laplacian Regularized Low-Rank Representation (TLR-LRR) algorithm by embedding a tensor hypergraph model in the data structure. Sections 4 and 5 present the experimental results and the conclusion, respectively.

## II. GENERAL FORMULATION

Consider a set of observations $Y \in \mathbb{R}^{M \times N}$ with $M$ being the dimensionality of the ambient space and $N$ the number of observations. This dataset is called "self-expressive" if its columns are possible to be linearly reconstructed from the other columns. In other words, observations can be considered as atoms of a dictionary to reconstruct other observations using the coefficient matrix $Z \in \mathbb{R}^{N \times N}$ and the reconstruction error matrix $E \in \mathbb{R}^{M \times N}$. Then, $Z$ is obtained by solving,

$$\min_{Z,E} rank(Z) + \gamma \|E\|_1 \quad s.t. \quad Y = YZ + E. \qquad (1)$$

As the rank function is non-convex, (1) can be relaxed by the nuclear norm as

$$min_{Z,E} \|Z\|_* + \gamma \|E\|_1 \quad s.t. \quad Y = YZ + E. \qquad (2)$$

To encourage reconstruction of each point from its close neighbour points in the Euclidean space, the constraint of sparsity over the coefficients matrix can also be added to the optimization problem [13]. Requiring the coefficients of $Z$ to be nonnegative, each data point can be reconstructed from a convex set of its surrounding points. Such constraints improve the subspace clustering accuracy based on the graph cut [27] of $Z$, as an adjacency matrix representing a graph in the projective space [22]. In this In this way, (2) is presented as

$$\min_{Z,E} \|Z\|_* + \lambda \|Z\|_1 + \gamma \|E\|_1,$$
$$s.t. \quad Y = YZ + E, \quad Z \geq 0. \qquad (3)$$

To preserve the local information around each data point, the regularized LRR methods suggest defining the weights as

$$[W]_{ij} = \begin{cases} 1; & if\ y_j \in N_k(y_i)\ or\ y_i \in N_k(y_j) \\ 0; & Otherwise, \end{cases} \qquad (4)$$

where $N_k(y_i)$ shows the set of $k$ nearest neighbours to each data point $y_i$ and $W \in \mathbb{R}^{N \times N}$ is the graph weight matrix satisfying the problem,

$$\min_{\{z_k\}} \sum_{ij} [W]_{ij} \|z_i - z_j\|^2, \quad (5)$$

where $z_i$ and $z_j$ are the projections of $y_i$ and $y_j$ onto the low-rank space. The above term can be equivalently presented as

$$\min tr(ZLZ^T), \quad (6)$$

which is incorporated as a regularization term in (3) to have,

$$\min_{Z,E} \|Z\|_* + \lambda \|Z\|_1 + \beta tr(ZLZ^T) + \gamma \|E\|_1, \quad (7)$$
$$s.t. \quad Y = YZ + E, \quad Z \geq 0.$$

In [13], it is suggested to embed a hypergraph $\mathcal{H} = (\mathcal{V}, \mathcal{E})$, in the ambient space based on observations [24], where $\mathcal{V}$ and $\mathcal{E}$ show a vertex and hyperedge set, respectively. Then, for each $e \in \mathcal{E}$, there is a set of vertices linked by the corresponding hyperedge weight $w(e)$, which is normally nonnegative. The degree of each vertex is defined as $d(v) = \sum_{e \in \mathcal{E}} w(e) h(v, e)$, where $h(v, e)$ is an entry of the matrix $H$ given by

$$h(v, e) = \begin{cases} 1 & if v \in e \\ 0 & otherwise. \end{cases} \quad (8)$$

Accordingly, to minimize the distance between the projection of all vertices of one hyperedge, [13] modifies (7) as

$$\min_{Z,E} \|Z\|_* + \lambda \|Z\|_1$$
$$+ \beta \sum_{i,j \in e \in \mathcal{E}} \frac{w(e)}{d(e)} \|z_i - z_j\|_2^2 + \gamma \|E\|_1, \quad (9)$$
$$s.t. \quad Y = YZ + E, \quad Z \geq 0.$$

Next, by assuming a homogeneous weight for the pairwise links in one hyperedge and after some manipulation, the above problem is expressed as

$$\min_{Z,E} \|Z\|_* + \lambda \|Z\|_1 + \beta tr(ZL^hZ^T) + \gamma \|E\|_1 \quad (10)$$
$$s.t. \quad Y = YZ + E, Z \geq 0$$

where,

$$L^h = D_\mathcal{V} - HW_\mathcal{E} D_\mathcal{E}^{-1} H^T, \quad (11)$$

with $W_\mathcal{E}$ showing the matrix of hyperedge weights, $D_\mathcal{V}$ being a diagonal matrix whose entries present the degree of vertices, and $D_\mathcal{E}^{-1}$ denoting the diagonal matrix of hyperedge degrees given by $d(v) = \sum_{e \in \mathcal{E}} w(e) h(v, e)$ [13].

## III. PROPOSED TLR-LRR ALGORITHM

The notion of using hypergraphs in order to present complex relationships between observations instead of only pairwise relationships has proven to be efficient in capturing the locality information in data points. However, the hypergraph method addressed in [13] has a major drawback. It inevitably poses a *fixed number of neighbours* around each observation. This may cause the algorithm to fail to capture the locality information effectively when the density of the observations is not uniform. Moreover, it applies a binary scheme of weighing for hyperedges, from which no information of variations in the observation density is recorded in the ambient space. Furthermore, as severe nonlinearities in the data structure can lead to major differences between the Euclidean and geodesic distance, these deficiencies can widely degrade the clustering performance. This becomes even worse in presence of outliers, as they might be taken a member of a neighborhood due to the posed fixed number of neighbors into the optimization problem.

In this paper, we propose a hypergraph for data modelling, which allows having a varied number of neighbors regarding the density of data points. This requires a new definition of adjacency for establishing hypergraphs and a new mathematical modelling for preserving the hyperedge weights through the projection onto the low-rank space. To proceed, the following preliminaries are required.

*A. Preliminaries*

For each hyperedge $e_i \in \mathcal{E}$ of hypergraph $\mathcal{H} = (\mathcal{V}, \mathcal{E})$, the degree $|e_i|$ defines the number of vertices of the set $e_i$. Then, the maximum cardinality of the hypergraph is defined as

$$P = m.c.e(\mathcal{H}) = max\{|e_i|: e_i \in \mathcal{E}\}. \quad (12)$$

We propose to link each vertex by virtue of $\varepsilon$-ball adjacency [25] to all the other vertices with a distance less than $\varepsilon$ in the Euclidean space. This is different from the kNN definition used in [13]. In this way, as the unnecessary links among distant observations are removed, the locality information can be better captured. In other words, the information of the data density would not be lost during the mapping to the low-rank space.

The hypergraph is then represented with an adjacency tensor $\mathbf{A} \in \mathbb{R}^{N \times N \times \ldots \times N \ P \ times}$ defined as [26]

$$\mathbf{A} = a_{i_1, i_2, \ldots, i_c}, 1 \leq i_1, i_2, \ldots, i_c \leq N. \quad (13)$$

Supposing each hyperedge $e_l = \{v_{l1}, v_{l2}, \ldots, v_{lc}\} \in \mathcal{E}$ includes $c \leq p$ vertices, then, $e_i$ is represented by all the entries $a_{i_1, i_2, \ldots, i_c}$ of $\mathbf{A}$, where a subset of c indices from $\{p_1, p_2, \ldots, p_c\}$ are not different from $\{l_1, l_2, \ldots, l_c\}$ in order to preserve the symmetry of $\mathbf{A}$. The other $p - c$ indices are selected randomly. More formally, the nonzero entries are selected as

$$s_{p_1 \ldots p_P} = u\left(\sum_{k_1, k_2, \ldots, k_c \geq 1, \sum_{i=1}^c k_i = P} \frac{P!}{k_1! k_2! \ldots k_c!}\right). \quad (14)$$

in which $u(.): \mathbb{R} \to \mathbb{R}$ shows the Heaviside function. The absolute value of nonzero entries can be a constant or equal to $c - 1$ as suggested in [26] or proportional to the density of observations within an $\varepsilon$-ball as follows,

$$a(s_{p_1 \ldots p_P}) = \frac{1}{c} \left(\sum_{i,j \in \{p_1, \ldots, p_P\}} \|y_i - y_j\|_2^2\right)^{-1} \quad (15)$$

By forming the weight tensor $\mathbf{A} \in \mathbb{R}^{N \times N \times \ldots \times N(P \ times)}$, the superdiagonal tensor of the degrees $\mathbf{D}$ is also of the same dimensionality with each diagonal entry giving the vertex degrees,

$$d_{ii \ldots i(p \ times)} = d(v_i) = \sum_{\{e_i: v_i \in e_i\}} a(e_i). \quad (16)$$

The tensor Laplacian of the hypergraph $\mathcal{H}$ is then given by

$$\mathbf{L} = \mathbf{D} - \mathbf{A} \quad (17)$$

The proposed optimization problem is derived in the following.

*B. Optimization Problem Definition*

Aiming at capturing the locality information, it is desired to minimize the distance between the adjacent data points in the ambient space by mapping to the latent space. Using the $\varepsilon$-ball adjacency definition, this is formally written as

$$\min_{\{z_k\}} \sum_{l1, l2, \ldots, lM = 1}^{N} \sum_{ij} \|z_i - z_j\|^2$$
$$\times a_{(all \ combinations \ of \ indices \ including \ i \ and \ j)} \quad (18)$$

in which the index combinations are given by (15). By some manipulations, (18) can also be presented as

$$\min_{\{z_k\}} \sum_{l1,l2,...,lM=1}^{N} \sum_{ij}(z_i^T z_j)^2 \quad (19)$$
$$\times a_{(all\ combinations\ of\ indices\ including\ i\ and\ j)},$$

or equivalently in compact form as
$$\min tr(Z \times_N \mathbf{L} \times_N Z^T), \quad (20)$$

where $\mathbf{L}$ is defined by (17), $\times_N$ is an *N*-fold matrix-tensor product [26], and the $trace(.)$ is performed on the main diagonal of the tensor $(Z \times_n \mathbf{L} \times_n Z^T)$ with *N* dimensions and *P* modes. Now, we add (20) to the LRR problem with non-negativity and sparsity constraint as a regularization term to complete the proposed optimization problem as

$$\min_{Z,E} \|Z\|_* + \lambda\|Z\|_1 + \beta tr(Z \times_n \mathbf{L} \times_n Z^T) + \gamma\|E\|_1$$
$$s.t.\ Y = YZ + E, Z \geq 0. \quad (21)$$

### C. Optimization Solution

Problem (21) can be solved using the Linearized Adaptive Direction of Multipliers Method (ADMM) [23] by first introducing an auxiliary variable $J \in \mathbb{R}^{N \times N}$ as,

$$\min_{J,Z,E} \|J\|_* + \lambda\|Z\|_1 + \beta tr(Z \times_n \mathbf{L} \times_n Z^T)$$
$$+ \gamma\|E\|_1 \quad (22)$$
$$s.t.\ Y = YZ + E, Z = J, J \geq 0.$$

Then, the augmented Lagrangian function for (22) is given by

$$\mathcal{L}(J,Z,E,M_1,M_2) = \|Z\|_* + \lambda\|J\|_1$$
$$+ \beta tr(Z \times_n \mathbf{L} \times_n Z^T)$$
$$+ \gamma\|E\|_1 + \langle M_1, Y - YZ - E\rangle$$
$$+ \langle M_2, Z - J\rangle \quad (23)$$
$$+ \frac{\mu}{2}(\|Y - YZ - E\|_F^2$$
$$+ \|Z - J\|_F^2),$$

where $\langle\cdot\rangle$ denotes the inner product and $M_1$, $M_2$, and $\mu$ are the Lagrangian multipliers. Assuming that all the variables except one are fixed, updating $Z$ is equivalent to minimizing $\mathfrak{L}_1$ defined as

$$\mathfrak{L}_1 = \|Z\|_* + \beta tr(Z \times_n \mathbf{L} \times_n Z^T) + \frac{\mu}{2}\left\|Y - YZ - E_k + \frac{1}{\mu}M_1^k\right\|_F^2 + \frac{\mu}{2}\left\|Z - J_k + \frac{1}{\mu}M_2^k\right\|_F^2, \quad (24)$$

which cannot be solved in closed form. Thus, to facilitate arriving at a closed form solution, the smooth part of $\mathfrak{L}_1$ given by

$$q(Z,J_k,E_k,M_1^k,M_2^k) = \beta tr(Z \times_n \mathbf{L} \times_n Z^T) + \frac{\mu}{2}\left\|Y - YZ - E_k + \frac{1}{\mu}M_1^k\right\|_F^2 + \frac{\mu}{2}\left\|Z - J_k + \frac{1}{\mu}M_2^k\right\|_F^2, \quad (25)$$

can be replaced by the two last terms of (26) as

$$\min_Z \|Z\|_* + \langle\nabla_Z q(Z_k), Z - Z_k\rangle + \frac{\eta_1}{2}\|Z - Z_k\|_F^2 \quad (26)$$

where $\nabla_Z q(Z_k)$ is the derivative of $q(Z,J_k,E_k,M_1^k,M_2^k)$ with respect to $Z$ at the $k_{th}$ iteration of the recursive updates, *i.e.* $Z_k$. The closed form update equation for $Z$ is then obtained as

$$Z_{k+1}^* = \Theta_{(\eta_1)^{-1}}\left(\frac{Z_k - \nabla_Z q(Z_k)}{\eta_1}\right), \quad (27)$$

where the asterisk is the optimal solution at the $k_{th}$ iteration, and $\Theta_\varepsilon(A) = US_\varepsilon(\Sigma)V^T$ is the Singular Value Thresholding (SVT) [27] for an arbitrary matrix $A = U\Sigma V^T$, and $S_\varepsilon(x)$ is defined as

$$S_\varepsilon(x) = sgn(x)max(|x| - \varepsilon, 0). \quad (28)$$

In [27], it is shown that the linear approximation of (25) is valid as long as $\eta_1 > 2\beta\|L\|_2 + \mu(1 + \|Y\|_2^2)$.

Updating $E$ can be performed in a completely independent procedure by solving the following minimization problem,

$$\min_E \gamma\|E\|_1 + \frac{\mu}{2}\left\|Y - YZ - E_k + \frac{1}{\mu}M_1^k\right\|_F^2, \quad (29)$$

which leads to the closed form solution,

$$E_{k+1} = S_{\frac{\gamma}{\mu}}\left(Y - YZ + \frac{1}{\mu}M_1^k\right). \quad (30)$$

In a similar independent fashion, updating $J$ is carried out by solving the problem:

$$\min_{J \geq 0} \lambda\|J\|_1 + \frac{\mu}{2}\left\|J - \left(Z + \frac{1}{\mu}M_2\right)\right\|_F^2, \quad (31)$$

which results in a closed-form solution given by

$$J_{k+1} = \max\{S_{\frac{\lambda}{\mu}}\left(Z_{k+1} + \frac{1}{\mu}M_2^k\right), 0\}. \quad (32)$$

In order to accelerate the algorithm convergence as a function of the step size, the Lagrangian multipliers can also be arbitrarily updated as [23]

$$M_1^{k+1} = M_1^k + \mu_k(Y - YZ_{k+1} - E_{k+1}) \quad (33)$$
$$M_2^{k+1} = M_2^k + \mu_k(Z_{k+1} - J_{k+1}) \quad (34)$$
$$\mu_{k+1} = \min(\mu_{max}, \rho_k\mu_k), \quad (35)$$

where $\rho_k$ is updated by

$$\rho_k = \begin{cases} \rho_0, & if\ max\{h_1,h_2,h_3\} \leq \varepsilon_2 \\ 1 & otherwise \end{cases}, \quad (36)$$

in which $h_1 = \eta_1\|Z_{k+1} - Z_k\|$, $h_2 = \mu_k\|J_{k+1} - J_k\|$, and $h_3 = \mu_k\|E_{k+1} - E_k\|$. According to [23], the convergence is achieved if the following inequalities are satisfied,

$$\frac{\|Y - YZ_{k+1} - E_{k+1}\|}{\|Y\|} < \varepsilon_1, \quad (37)$$
$$max\{h_1, h_2, h_3\} \leq \varepsilon_2. \quad (38)$$

It is noteworthy to mention that the convergence of the proposed algorithm is guaranteed as long as the Lagrangian multipliers are set according to the aforementioned setting.

Algorithm 1 gives a summary of the TLR-LRR algorithm with a computational complexity of $\mathcal{O}(MN^2)$.

**Algorithm 1.** Tensor Laplacian Regularized LRR (Proposed)

---
**Input:** $Y \in \mathbb{R}^{M \times N}$,
$\lambda = 0.01, \beta = 10, \gamma = 1.1, \varepsilon_1 = 10^{-6}, \varepsilon_2 = 10^{-4}$ and $\epsilon = 0.05$
**Output:** $Z \in \mathbb{R}^{N \times N}, E \in \mathbb{R}^{M \times N}$
**Initialization:** Compute P the maximum number of neighbours given by the adjacency limit
**Compute** $\mathbf{L} \in \mathbb{R}^{N \times N \times ... \times N(P\ times)}$
$Z_0 = J_0 = M_1^0 = 0, E_0 = M_2^0 = 0$, k=0
**While** (not converged) repeat:
  1: $k := k + 1$
  2: Update $Z$ using (28)
  3: Update $E$ using (30)
  4: Update $J$ using (32)
  5: Update Lagrangian Multipliers using (33)-(35)
  6: Check convergence under the conditions (37) and (38)
**End while**.

## IV. EXPERMENTAL RESULTS

The TLR-LRR algorithm is evaluated in subspace clustering for both synthetic and real-world datasets. The clustering accuracy is measured by

$$AC = \frac{\sum_{i=1}^{N} \delta(\hat{\mathcal{F}}(i), Match(\hat{\mathcal{F}}, \mathcal{F}))}{N}, \quad (39)$$

where $N$ is the total number of predictions, $\delta(a, b)$ is equal to 1, only if $a = b$, and $Match(.)$ is the best matching function that permutes $\hat{\mathcal{F}}$ to match with $\mathcal{F}$ to fulfill the Kuhn-Munkres algorithm [28]. This means that the assigned label to the clusters may be permuted while solely considering the distinction between them. In the following, we evaluate the performance of the TLR-LRR algorithm in data clustering for both synthetic and real data.

### A. Synthetic data

The synthetic datasets are generated in two dimensions. To observe the results under different conditions, the data exploit various features such as overlapping among the clusters and variation in the data density. Also, to show the resilience of the proposed algorithm against changes in data structure, we consider both linearly independent and correlated clusters.

Fig. 1 shows the ground truth clustering synthetic datasets. The results are compared with the state-of-art methods: Laplacian Regularized Low-Rank Representation (LRLRR) [13] and Adaptive Laplacian Low-Rank Representation (ALLRR) [22]. Figs. 2 and 3 illustrate the results by applying the Normalized Graph Cut (NCut) [29] on Z as an adjacency matrix in an $N$-dimensional space. Table 1 reports the clustering results for *k*-means [28], NCut [27], basic LRR [6], ALLRR [22], LLRR [13], and TLR-LRR algorithms.

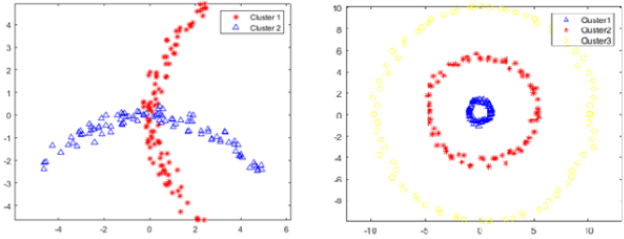

Fig. 1. Ground truth synthetic datasets; two-moon (left) and three circles (right).

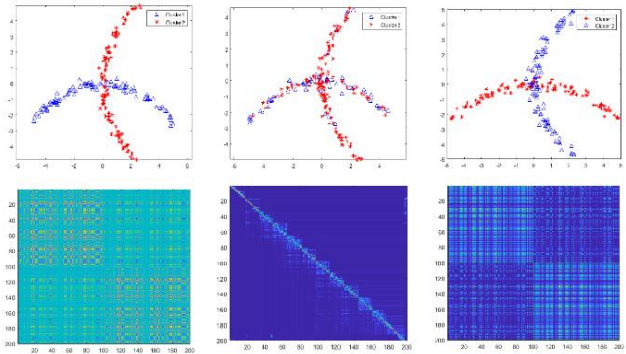

Fig. 2. Two moons clustering results (top row) and the corresponding coefficient matrices (bottom below); ALLRR (left), LRLRR (middle), TLR-LRR (right).

Table 1. Clustering accuracies for synthetic datasets.

|  | Two moons | Three circles |
|---|---|---|
| K-Means | 0.54 | 0.33 |
| NCut | 0.58 | 0.41 |
| Simple LRR | 0.61 | 0.45 |
| ALLRR | 0.79 | 0.93 |
| LRLRR | 0.94 | 0.45 |
| **TLR-LRR (Proposed)** | **0.99** | **0.98** |

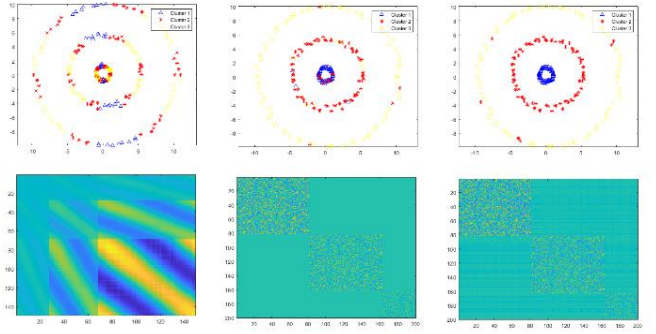

Fig. 3. Three circles clustering results (top row) and corresponding coefficient matrices (bottom row) ALLRR (left), LRLRR (middle), TLR-LRR (right).

### B. Real Data

Image clustering is performed on the datasets: AT&T [35], Extended YaleB [36], and USPS [37]. The Extended YaleB is a famous facial image dataset with 16128 images of 28 people with 9 different gestures and under 64 different lightening conditions. The USPS is a dataset of 20,000 images of hand-written digits. The clustering accuracies are presented in Table 2. From the results, one can observe that by incorporation of the locality information on the manifold structure, the proposed TLR-LRR algorithm achieves higher clustering accuracy for both synthetic and real datasets. Table 3 compares the computational complexity of the proposed method with other benchmarks.

Table 2. Clustering accuracies for real datasets.

|  | AT&T | Extended YaleB | USPS |
|---|---|---|---|
| *k*-means | 0.12 | 0.37 | 0.23 |
| NCut | 0.27 | 0.21 | 0.45 |
| Simple LRR | 0.35 | 0.32 | 0.45 |
| ALLRR | 0.54 | 0.48 | 0.47 |
| LRLRR | 0.65 | 0.88 | 0.89 |
| **TLR-LRR (Proposed)** | **0.78** | **0.92** | **0.97** |

Table 3. Computational complexity of the algorithms

| Algorithm | ALLRR | LRLRR | **TLR-LRR (Proposed)** |
|---|---|---|---|
| Complexity order | $\mathcal{O}(KMN^2)$ | $\mathcal{O}(MN^2)$ | $\mathcal{O}(MN^2)$ |

## V. CONCLUSION

We proposed the TLR-LRR algorithm for clustering the observations with varied densities in manifold structures, which was implemented by incorporation of hypergraph modeling. Moreover, the sparsity of subspaces was considered to encourage the linear reconstruction of each node using its adjacent data points in the Euclidean space. In this way, the resilience of the method was improved against nonlinearities and correlation between the sub-components of the geometrical structure of the underlying data. Additionally, a low-rank and sparse coefficients matrix was constrained to be non-negative in order to place each reconstructed node within a convex set of other observations for further resilience against outliers. The optimization problem was solved using the ADMM, with the computational complexity of $\mathcal{O}(MN^2)$, similar to and less than the other two state-of-art methods. Simulation results showed that the proposed method achieves more accurate clustering results compared to the well-known methods for both synthetic and real datasets.